\documentclass[a4paper, fleqn]{cas-dc}

\usepackage{amsmath,amsfonts}
\usepackage{algorithmic}
\usepackage{graphicx}
\usepackage{textcomp}
\usepackage{url}
\usepackage{xcolor}
\usepackage{times}
\usepackage{epsfig}
\usepackage{amssymb}
\usepackage{listings}
\usepackage{booktabs} 
\usepackage{multirow}
\usepackage{gensymb}
\usepackage{mathtools}
\usepackage{adjustbox}
\usepackage{framed}
\usepackage{latexsym}
\usepackage{float}
\usepackage{placeins}
\usepackage{flafter}
\usepackage[numbers,sort&compress,longnamesfirst]{natbib}

\ExplSyntaxOn
\cs_gset:Npn \__first_footerline:
  { \group_begin: \small \sffamily \__short_authors: \group_end: }
\ExplSyntaxOff 
\def\tsc#1{\csdef{#1}{\textsc{\lowercase{#1}}\xspace}}
\tsc{WGM}
\tsc{QE}

\begin{document}
\let\WriteBookmarks\relax
\def\floatpagepagefraction{1}
\def\textpagefraction{.001}

\shorttitle{Detection of MCI Using Facial Features in Video Conversations}    

\shortauthors{Alsuhaibani \textit{et al.}}  

\title [mode = title]{Detection of Mild Cognitive Impairment Using Facial Features in Video Conversations}

\author[DU,sattam]{Muath Alsuhaibani}
\ead{muath.alsuhaibani@du.edu}
\credit{Conceptualization, Data curation, Formal analysis, Investigation, Methodology, Software, Validation, Visualization, Writing - original draft}

\author[harvard]{Hiroko H. Dodge}
\ead{hdodge@mgh.harvard.edu}
\credit{Investigation, Resources, Validation, Writing - review \& editing}

\author[DU]{Mohammad H. Mahoor}\corref{cor}
\ead{mohammad.mahoor@du.edu}
\credit{Investigation, Resources, Supervision, Project administration, Validation, Writing - review \& editing}

\affiliation[DU]{organization={Department of Electrical and Computer Engineering, University of Denver},
            city={Denver},
            citysep={}, 
            postcode={80208}, 
            state={CO},
            country={United States}}

\affiliation[sattam]{organization={Department of Electrical Engineering, Prince Sattam Bin Abdulaziz University},
            city={Al-Kharj},
            citysep={}, 
            postcode={11942}, 
            country={Saudi Arabia}}

\affiliation[harvard]{organization={Department of Neurology, Massachusetts General Hospital, Harvard Medical School},     
            city={Boston},
            citysep={}, 
            postcode={02114}, 
            state={MA},
            country={United States}}

\cortext[cor]{Corresponding author}

\begin{abstract}
Early detection of Mild Cognitive Impairment (MCI) leads to early interventions to slow the progression from MCI into dementia. Deep Learning (DL) algorithms could help achieve early non-invasive, low-cost detection of MCI. This paper presents the detection of MCI in older adults using DL models based only on facial features extracted from video-recorded conversations at home. We used the data collected from the I-CONECT behavioral intervention study (NCT02871921), where several sessions of semi-structured interviews between socially isolated older individuals and interviewers were video recorded. We develop a framework that extracts spatial holistic facial features using a convolutional autoencoder and temporal information using transformers. Our proposed DL model was able to detect the I-CONECT study participants' cognitive conditions (MCI vs. those with normal cognition (NC)) using facial features. The segments and sequence information of the facial features improved the prediction performance compared with the non-temporal features. The detection accuracy using this combined method reached 88\% whereas 84\% is the accuracy without applying the segments and sequences information of the facial features within a video on a certain theme.
\end{abstract}




\maketitle
\section{Introduction}

The Alzheimer's Association is projecting the number of Americans age 65 and older who are living with Alzheimer's disease-related dementia to double by 2050. Alzheimer's Disease (AD) is a brain disease affecting memory and other cognitive functions, ultimately deteriorating a patient's ability to perform daily tasks. Mild Cognitive Impairment (MCI) has a high likelihood to progress to AD overtime. Older adults (MCI) can live independently but they start having difficulty with tasks that depends on memory, language, or other thinking skills. Currently, older adults with MCI are diagnosed clinically; however, their daily challenges are often not noticeable to those whom they encounter irregularly. Early diagnosis of MCI helps to identify people who are at increased risk of developing dementia. Once identified, those with MCI can then be offered information, advice, and support and be kept under review. Also, if they do develop dementia, they would have been diagnosed sooner and receive treatment \cite{noauthor_2022_2022}.

It is often challenging to diagnose MCI as it requires in-depth neuropsychological assessments \cite{apostolova2008neuropsychiatric}, brain positron emission tomography (PET) scan, or spinal fluid test for amyloid beta protein \cite{herholz2007positron, rabinovici2019association, herukka2017recommendations, butterfield2007roles, ritchie2014plasma}. Although recent advancements in plasma blood biomarkers could provide more affordable solutions in the future, we need to come up with cost-effective, non-invasive approaches to identify MCI at an early stage in the community.

Artificial Intelligence (AI) has shown to be a promising technique for achieving an automated detection of cognitive impairment conditions at an early stage. Thus, the number of research in recent years that aim to use AI in medical condition detection has surged. AI showed that extracting valuable features from users' physiological, biological, and behavioral patterns would detect the human medical condition. Although many AI studies are focused on clinical assessments and medical scans such as PET, magnetic resonance imaging (MRI), functional MRI (fMRI), and electroencephalogram (EEG) \cite{jo2020deep, kam2019deep, yang2019m}, further research needs to be developed in analyzing behavioral patterns using Deep Learning (DL) methods. 

Deep learning algorithms are an essential subset of AI. The DL models include various architectures that have been proven to resolve substantial challenges in many fields including computer vision. The DL models help to extract useful features from large amounts of data. The outstanding achievements of AI systems in digital healthcare performance were achieved by deep learning. For example, several types of cancers are detected using DL models \cite{hu2018deep,shen2019deep,painuli2022recent}. 

Detection of medical conditions such as depression \cite{jiang2020classifying,fang2023multimodal}, Parkinson's \cite{bandini2017analysis}, or other psychiatric impairments \cite{birnbaum2022acoustic} has been successfully studied with patients' facial features using DL approaches. These algorithms are applied to analyze facial features. In this study, we capture facial features from semi-structured video interviews of MCI and normal cognition (NC) participants who underwent an I-CONECT clinical trial (clinicaltrials.gov\#: NCT02871921) \cite{yu2021internet}. This clinical trial was aimed to enhance cognitive reserve by providing frequent conversations to socially isolated older subjects. The interaction between the elderly and the study interviewers via a webcam could show the cognitive status of the older participants. We hypothesize that patterns in facial features among MCI make them distinguishable from those with normal cognition. We propose a DL method that detects cognitive conditions in older adults using their facial features in the I-CONECT dataset. The purpose of this paper is to show that the facial features and interaction information of the participants during the semi-structured interviews are impacted by the cognitive conditions of participants.

The remainder of this paper is structured as follows. Section \ref{sec.relatedwork} reviews the related work of detecting MCI using facial features. Section \ref{sec.materials_methods} describes the dataset and explains the preprocessing procedure and feature extraction. Section \ref{sec.experiments_results} provides the experimental implementation and results. Finally, Section \ref{sec.conclusion} concludes this paper with a discussion and future research directions.
\begin{figure*}
    \centering
    \includegraphics[width=\linewidth]{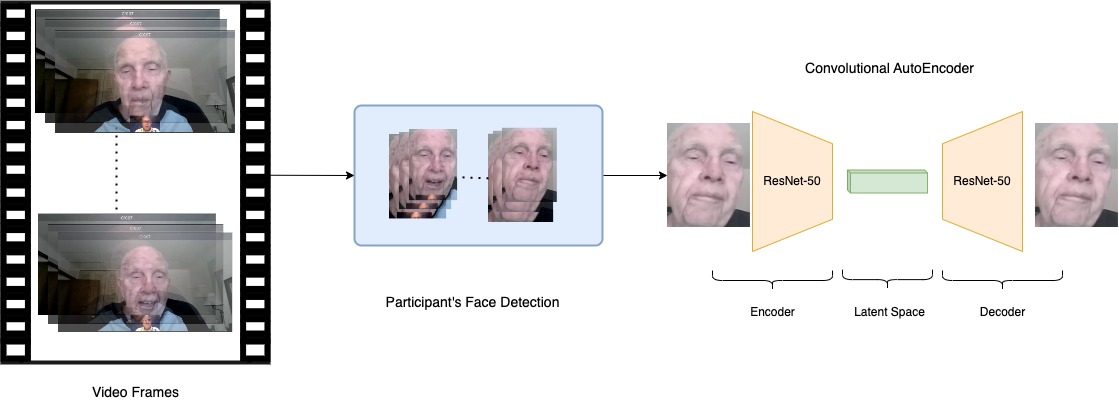}
    \caption{The Prepossessing Steps of Participants' Videos.}
    \label{fig.preprocessing}
\end{figure*}
\section{Related Work} \label{sec.relatedwork}
{
In this section, we will review past studies in the field of computer vision to determine human cognitions mainly based on facial features, including eye movements and facial expressions.

\textbf{Eye Features}: Several studies have investigated the differences in eye features between NC and MCI \cite{crawford2005inhibitory,chan2018eye,wilcockson2019abnormalities}. These eye features include eye movements, gaze, and blinking. Alzahrani \textit{et al.} \cite{alzahrani2021eye} computed an eye aspect ratio from six points of participants’ facial landmarks extracted using Dlib \cite{huang2008labeled} and Openface \cite{baltrusaitis2018openface} libraries, which have pre-trained models for facial features extraction. They calculated the eye blinking rate for every participant and applied basic machine learning algorithms to predict participants' cognitive conditions. Because they used two different pre-trained models, they reported a 10\% accuracy difference between the models.
Chen \textit{et al.} \cite{chan2018eye} have analyzed the eye movement patterns of participants while performing face recognition tasks. They found that older adults tend to focus on the center of the target face especially the elderly with low Montreal Cognitive Assessment (MoCA) scores. 
Furthermore, Nam \textit{et al.}\cite{nam2020analyzing} recorded participants' faces while they watch emotion-evoking videos to analyze the correlation between the eye and head movements. Using Openface \cite{baltrusaitis2018openface}, they extracted the eye and head movements. They analyzed these data in order to observe the attention and concentration decline of dementia patients. They concluded that AD patients tend to have eye movements in the same vertical direction as their heads. Haque \textit{et al.} \cite{haque2020deep} developed a deep learning model which estimates participants’ eye gaze during memory-triggering tasks in a clinical setting. They were able to distinguish cognitive impairment participants by tracking their eye gaze and viewing time during the tasks.

\textbf{Facial Expression}: Studies have shown that cognitive impairment does affect a person's facial expression \cite{mclellan2008recognition,spoletini2008facial}. Using computer vision approaches, researchers have extracted facial action units \cite{tanaka2019detecting} or estimated facial expression using deep learning methods \cite{fei2022novel} to predict participants' cognitive condition. 
Jiang \textit{et al.} \cite{jiang2022automated} recorded participants' faces while undergoing a memory test in a clinical setting. They extracted participants' facial expressions using a pre-trained Convolutional Neural Network (CNN) on a popular facial expression dataset that is not dedicated to elderly faces. The estimated facial expression for the video frames are average for subject-level facial expression. Traditional machine learning algorithms were implemented to classify cognitively impaired participants from cognitively unimpaired ones. They also considered the viewing time of the targeted region. Fei \textit{et al.} \cite{fei2022novel} extracted participants’ facial expressions when participants are watching emotional stimuli. The authors computed the emotional occurrence among cognitive conditions elderly. They selected time periods of the participants' emotional occurrence which has a large difference between MCI and NC. A traditional machine learning algorithm was used to classify participants' conditions. Gerlowska \textit{et al.} \cite{gerlowska2021facial} showed no differences in the frequency of emotions expressed in patients and healthy controls. An emotional functioning pattern among older adults was noticed despite cognitive dysfunctions. The analysis was based on mimicking 6 basic emotions that were shown on a screen for the subjects and manually selecting 10-15 most intense seconds. 
Tanaka \textit{et al.} \cite{tanaka2019detecting} extracted facial landmarks using Openface \cite{baltrusaitis2018openface} from participants while responding to an agent system. They tracked facial landmarks that represent facial action units during the participants' response to the agent. They developed an algorithm to extract talking segments using a sequence of facial landmarks that surround the mouth. Using this approach, they were able to classify participants with dementia using traditional machine learning methods. 

\textbf{Holistic Facial Features}: There are limited works that use holistic facial features to detect participants' cognitive conditions. Umeda-Kameyama \textit{et al.} \cite{umeda2021screening} implemented deep learning models to detect cognitive impairment conditions of participants' facial images. Consequently, spatial facial features are extracted for the training and evaluation of the model. They found that the model had better prediction performance when using the lower half of participants' facial images. Lee \textit{et al.} \cite{lee2022detection} applied two-stream ConvNet using spatial and motion features to classify participants' conditions with frame sequence intervals of 10 frames per segment. 

We focused on studies that adapted computer vision algorithms in facial feature extraction either implementing DL models that are trained on large datasets to capture these features or proposing a model that is solely for the purpose of detecting the participants' cognitive status. Regardless of the adapted methods, the participant's facial features led to the revealing of the cognitive conditions. In most of the works, participants were in a clinical setting in which the study staff had control over the room lighting and point view of participants' faces \cite{jiang2022automated, fei2022novel}. However, video recording of participants' faces in a home setting is very challenging to have an automated method to extract facial features. We have faced some challenges in this respect. Additional challenges will be also discussed later in this paper. 

In addition to the reviewed above, facial landmarks extracted using pre-trained models do not achieve highly accurate results on the elderly, especially older adults with cognitive impairment conditions \cite{asgarian2019limitations,taati2019algorithmic}. Therefore, the detection of facial landmarks still needs to be improved for the elderly. This motivated our approach where we extracted holistic facial features using Convolutional Autoencoder (CAE). 

}
\section{Materials and Methods} \label{sec.materials_methods}
In this section, we introduce the I-CONECT dataset and our approach to preprocessing the participants' appearance during the interviews. We also explain our method of spatial and temporal extractions of features. 
\subsection{Dataset} \label{sec.dataset}
{
\begin{table}[ht]
\caption{Themes general information.}
\label{theme_info_table}
\resizebox{\columnwidth}{!}{
\begin{tabular}{lccc}
\hline
Themes           & Trial's Week & Total Participants & Selected Videos \\ \midrule
Summertime       &  1& 66                 & 30               \\
Self-Care         &  2& 60                 & 30                  \\
Halloween        &  23& 59                 & 32                  \\
Cities and Towns &  9& 59                 & 39                \\\hline
\end{tabular}}
\end{table}
\begin{table}[ht]
\caption{Demographic information of participants for selected themes.}
\label{demog_table}
\resizebox{\columnwidth}{!}{
\begin{tabular}{lcccc}
\hline
Themes           & \begin{tabular}[c]{@{}c@{}}Age, Years\\  (mean) (SD)\end{tabular} & \begin{tabular}[c]{@{}c@{}}Education\\  (mean)  (SD)\end{tabular} & \begin{tabular}[c]{@{}c@{}}Condition\\  (MCI/NC)\end{tabular} & \begin{tabular}[c]{@{}c@{}}Gender\\  (M/F)\end{tabular} \\ \hline
Summertime       & 80.7  (4.75)                                                      & 15.8  (2.79)                                                      & 14/16                                                          & 8/22                                                    \\
Self-Care        & 80.5  (4.24)                                                      & 15.4  (2.63)                                                      & 16/14                                                          & 11/19                                                   \\
Halloween        & 80.3  (3.73)                                                      & 15.3  (2.5)                                                       & 14/18                                                          & 12/20                                                   \\
Cities and Towns & 80.7  (4.31)                                                      & 15.5  (2.59)                                                      & 19/20                                                          & 15/24                                                   \\ \hline
\end{tabular}}

\end{table}

The Internet-Based Conversational Engagement Clinical Trial (I-CONECT) has video-recorded semi-structured interviews between the socially-isolated-older participants and trained conversational staff \cite{yu2021internet}. The participants have a minimum age limit of 75 and live in socially isolated environments to be included in this study. These participants were clinically diagnosed with either MCI or NC at the baseline. For each session, the participants and interviewer have a specific theme to discuss such as pets, foods, history, etc. Each session is approximately 30 minutes in duration. A total of 158 themes of video chats have been recorded among 69 subjects who were assigned to the intervention (i.e. video-chats) group.

We have selected four themes for this study based on the number of participants with recordings. Table \ref{theme_info_table} shows the overview of selected themes (Summertime, Self-Care, Halloween, Cities and Towns) with the trial's week number, the total number of participants that attended, and the number of selected videos based on the quality evaluation which we discuss in Section \ref{seq.preprocessing}. Demographic characteristics of participants vary between every theme since some participants have missed these themes' sessions or their videos are excluded in the current study due to quality issues with the videos. Since the selected participants of the themes are inconsistent, Table \ref{demog_table} shows the demographics by the themes. 

In our study, we made sure that videos with the same topic were analyzed separately because topics would trigger specific emotions that do affect facial features. The participants were assessed for their cognitive status (i.e., MCI vs. NC) at three-time points (baseline, Month 6, Month 12). We used the cognitive status assessed closest to the video chats. In our experiment, we used the clinical status assessed at the beginning of the trial except for the Halloween theme which occurred near the six-month (24 weeks) evaluation. 
}

\subsection{Data preprocessing} \label{seq.preprocessing}
{
The I-CONECT dataset requires preprocessing to make it suitable for training on our model. The preprocessing steps include: selecting the participants' appearance during the videos, face extraction, quality evaluation, and facial feature extraction. The videos include the participants' IDs during the facial appearance. Thus, optical character recognition (OCR) helped to determine the desired frames from the video with less computation and higher confidence. We used CRAFT \cite{baek2019character} which is a pre-trained model that performs scene text detection on natural images. Because the videos have different frame rates, we fixed the frame extraction rate of the videos. We performed a downsampling of the original frame per second (fps) which ranged from 30 fps to 10 fps. In our experiment, we choose a frame rate of 10 fps by following the frameshift value from Equation \ref{eq.shiftstep} to perform frame selection.
\begin{equation} \label{eq.shiftstep}
FrameShift = Rounddown\left(\frac{fps_{original}}{fps_{target}}\right)
\end{equation}

In order to detect participants' faces, we used the retina face pre-trained model \cite{deng2020retinaface} to detect and extract the subject faces. Figure \ref{fig.preprocessing} shows the preprocessing steps to extract the facial features from the participant video interviews. To ensure that all extracted faces belong to participants, the detected faces are referenced by face size and coordination within the frames. Faces within the created area of interest are kept which results in better face extraction of the participants. 

We randomly evaluated the majority of the extracted faces from every video. The quality evaluation criteria are as follows: \underline{1. Very good}: all the faces are clear and the lighting is good. \underline{2. Good}: all the faces are clear and the lighting is acceptable. \underline{3. Ok}: most of the faces are clear or need further processing to enhance the extracted faces. \underline{4. Poor}: some of the extracted faces are unclear, i.e. screen obstacle covers part of the faces. \underline{5. Very poor}: most of the extracted faces are not acceptable due to lighting conditions or subjects' faces being barely detectable. This step is performed manually because it is challenging to automate this step. The participants' videos that do not meet the quality standards are dropped. We set the passing limit for video ratings of very good and good.
}
\begin{table}[ht]
    \centering
    \caption{The ResNet-50 Architecture of 96 image size input.}
\begin{tabular}{ccc}
\hline
layer name                & output size       & 50-layer                                                                \\ \midrule
conv1                     & $48 \times 48$           & $7\times7, 64,$ stride 2                                         \\ 
\multirow{4}{*}{conv2\_x} & \multirow{4}{*}{$24\times24$} & $3\times3$ max pool, stride 2  \\ & &  $\begin{bmatrix}1\times1, & 64\\3\times3, & 64\\1\times1, & 256\end{bmatrix} \times 3 $              \\ 

conv3\_x                &       $12\times12$         & $\begin{bmatrix} 1\times1, & 128 \\ 3\times3, & 128 \\ 1\times1, & 512 \end{bmatrix} \times 4 $  \\
conv4\_x                  &     $6\times6$          & $\begin{bmatrix}1\times1, & 256\\3\times3, & 256\\1\times1, & 1024\end{bmatrix} \times 6 $  \\ 
conv5\_x                  &     $3\times3$              & $\begin{bmatrix}1\times1, & 512\\3\times3, & 512\\1\times1, & 2048\end{bmatrix} \times 3 $  \\ 
                          &                   & average pool, 128-d fc, softmax                                           \\ \hline
\end{tabular}

\label{table.resnet50}
\end{table}

\subsection{Unsupervised learning} 

\textbf{Facial Features}: Since there are no pre-trained CNN models suitable for the elderly especially those with cognitive impairment \cite{taati2019algorithmic}, we initialized an autoencoder neural network architecture which is integrated to solve the issue of extracting meaningful features from unlabeled data. The autoencoder depends on three components which are the encoder and decoder, and latent code. We integrated a CNN for the encoder and decoder of the autoencoder to extract subjects' facial features. We extracted the subject's facial features at the frame level. We applied a CAE neural network with ResNet-50 architecture \cite{he2016deep} in the encoder and the reverse architecture of ResNet-50 as a decoder with a latent feature size of 128. Table \ref{table.resnet50} shows the architecture of ResNet-50 with the output size of network layers based on our input size of the faces. Using this approach, we generated a feature vector of every frame. The feature vectors should have embedded features of facial expression, head pose, eye gaze, and general facial features. extracted faces are inconsistent in terms of width and height. Thus, they are resized to a width and height of 96. The face size is set to avoid upsampling of participants' faces.
\begin{figure}
    \centering
    \includegraphics[width=\linewidth]{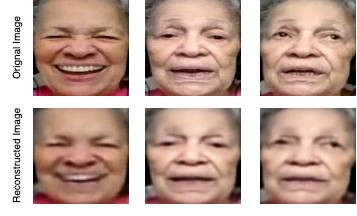}
    \caption{Examples of participant's face and autoencoder reconstructed Ones.}
    \label{fig.ae_image}
\end{figure}

\textbf{Feature analysis}: It is challenging to interpret the latent features of an autoencoder with the original image. For example, the values of the vector elements that represent facial expressions, poses, or eye gazes are still ambiguous. In fact, it is an ongoing research topic, nevertheless, we show that the reconstructed image preserved the facial features of the original image. Figure \ref{fig.ae_image} shows examples of a participant's face with different facial features that are encoded and decoded using our CAE model after training. Since we are dealing with videos, consecutive frames should have similar features with only slight changes, based on the participant's facial movement, which could represent the changes in facial expression and eye features as well.

\begin{figure}
    \centering
    \includegraphics[width=\linewidth]{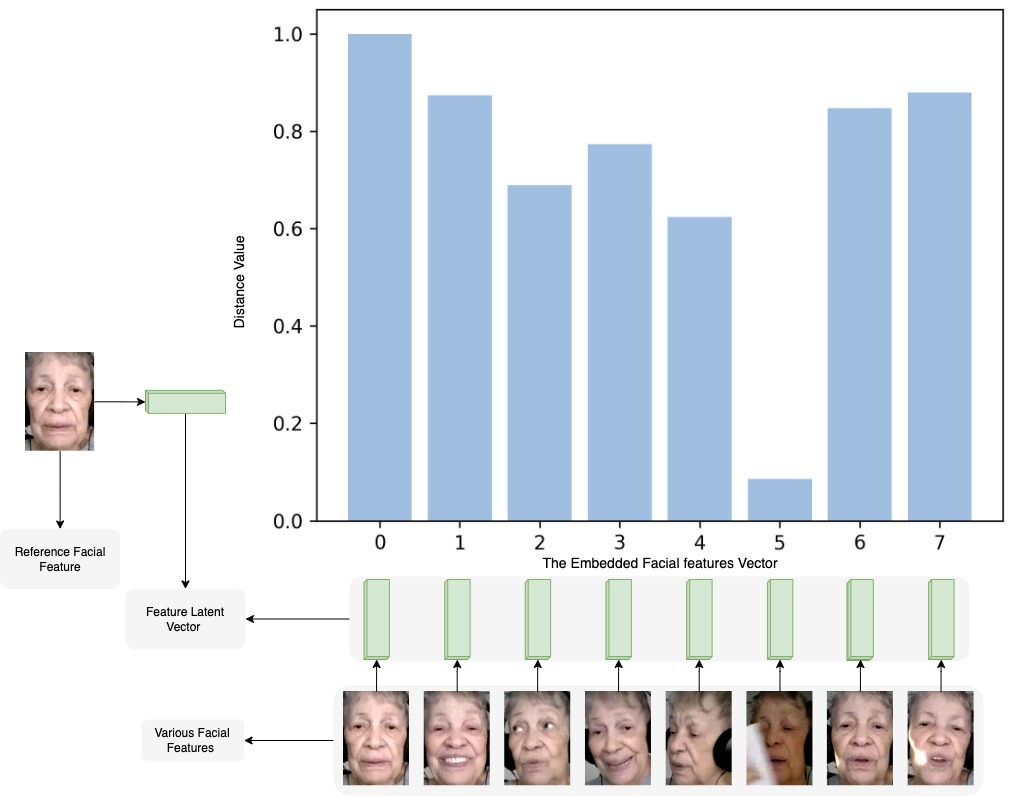}
    \caption{The vector cosine similarities between several latent feature vectors.}
    \label{fig.vector_similarity}
\end{figure}

The latent feature vectors have embedded participants' facial features based on their interaction during the interviews. The CAE preserves the related facial attributes in the latent feature vectors. In order to prove and ensure the relativity of facial features within these vectors, we have calculated the cosine similarity among various facial representations of the embedding features of the same subject. Figure \ref{fig.vector_similarity} shows the values of latent feature vector similarity among the selected facial frames of the same participant during the same theme discussion. We set a reference facial embedding vector that has a neutral facial expression, clear face presentation, and the participant's head pose is relatively ideal during the interview. This empirical result shows the difference in the facial feature representation. It is clear to observe that the vector that has the largest difference is when the participant raised a paper that covered part of the face which threw off the embedding features. Although the cosine similarity ranges from -1 to 1 where 1 is the same vector and -1 is the opposite vector, all these similarity have values above 0 which indicate they are similar in the facial features of the same participant.

\subsection{Temporal information}

\begin{figure*}
    \centering
    \includegraphics[width=\linewidth]{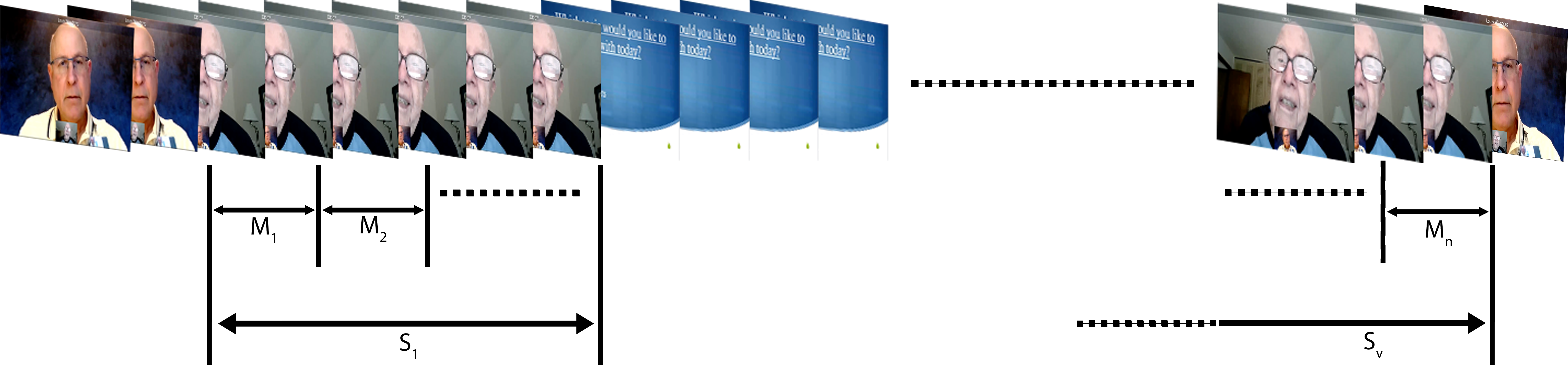}
    \caption{The frame extraction from a video with segments and sequences labeling.}
    \label{fig.seg_seq_labels}
\end{figure*}

\begin{figure}
    \centering
    \includegraphics[width=\linewidth]{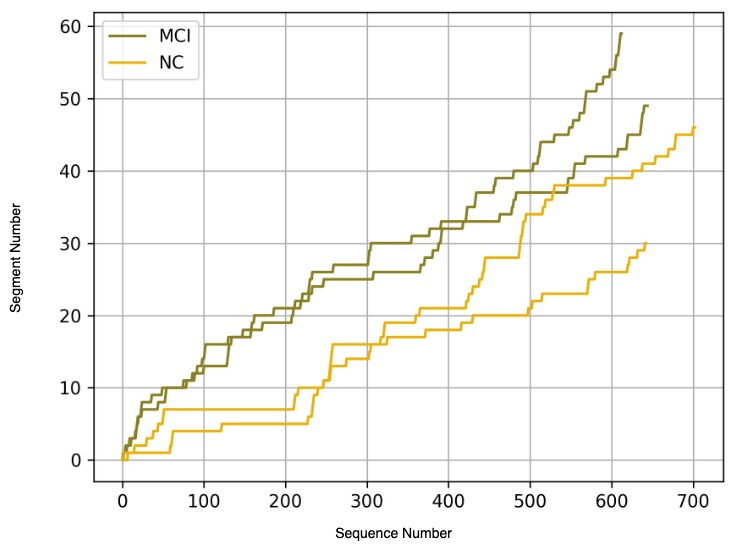}
    \caption{An example of a number of segments and sequences of participants with different cognitive conditions.}
    \label{fig.seg_seq_example}
\end{figure}

\textbf{Segments and sequences}: We have extracted features that fed to the model which represent segments and sequences of the participants' interaction. We define a segment as consecutive frames within a video that have extracted faces. A segment within a video is ended when three consecutive frames do not show a participant’s face as the main face in the frame. We also refer to a sequence as a set of consecutive frames. In the selection of the sequence and segments of the videos, a sequence must present all its interval facial features. Thus, the segment is dropped if its number of frames is less than the sequence size. During the feature arrangement, we made sure that consecutive frames are set to the same sequence. The video has three main parts which are the participant's face, the interviewer's face, and the theme introduction slides. We are interested in the participants' faces; thus, we kept the participants' frames and labeled their appearance as it is shown in Figure \ref{fig.seg_seq_labels}. Consequently, we are also showing a snippet of the number of segments and sequences of participants that have different cognitive conditions. Figure \ref{fig.seg_seq_example} shows the plot of segments and sequence indices within a video. Although this is a small example of the sequence and segment information, the figure leads to the idea that MCI participants have less continuous taking time compared to the NC participants. This indicates that interviewers tend to follow up more.

\textbf{Temporal features}: The transformers were introduced by \cite{vaswani2017attention} for machine translation. Although transformers are widely used in the Natural Language Processing field, the computer vision field also benefited from the concept and is being applied to images and videos \cite{dosovitskiy2020image,arnab2021vivit}. The idea of representing a frame as a token should help the prediction of subjects' conditions. Since transformers have proven their capability in various deep-learning interventions, they can capture temporal information from sequential frame features due to the attention mechanism represented in Equation \ref{eq.att} where Q, K, and V are query, key, and value, respectively. Transformers also use multi-head attention to perform a parallel computation of the attention mechanism where a key, query, and value are split into a number of heads and concatenated after the self-attention is applied. Equation \ref{eq.mha} shows the computation process of attention scores where self-attention is calculated for every head where $W^o$ is a learnable parameter that is multiplied by the concatenation of all the heads. Heads equation is shown in Equation \ref{eq.head}.
\begin{equation} \label{eq.att} 
Attention(Q,K,V) = softmax\left(\frac{QK^T}{\sqrt{d_k}}\right)V
\end{equation}
\begin{equation} \label{eq.mha} 
MultiHead(Q,K,V) = Concat(head_1, head_2, ... head_i)W^o 
\end{equation}
Where \begin{equation} \label{eq.head} 
head_i = Attention(QW_i^Q,KW_i^K,VW_i^V)
\end{equation}

Transformers are able to capture the dependency of the input data, so the representation of the feature vectors and classification tokens affect the behavior of the model. Thus, we created a classification token for every sequence which is a common practice for sequence classification using transformers \cite{devlin2018bert,dosovitskiy2020image}. The classification token is fed to a classifier to perform sequence classification of the input. Thus, this arbitrary token would have learned from the self-attention mechanism and positional embedding layer the label of the sequence input.

One of the important components of transformers is the positional embedding layer. This layer preserves the positions of the sequential representation for the attention mechanism. This layer inspired us to have the sequence and segment representation of every video added to the transformer. This positioning of the facial features could yield a hierarchical representation of an overall representation of a participant's video.

\section{Experiments and Results} \label{sec.experiments_results}
{
In this section, we present our experiment implementation details. Then, we explain the section performance of our model. Eventually, we study the model performance while changing some of the hyperparameters of the model in the ablation study.

\subsection{Implementation}
\begin{figure*}
    \centering
    \includegraphics[width=\textwidth]{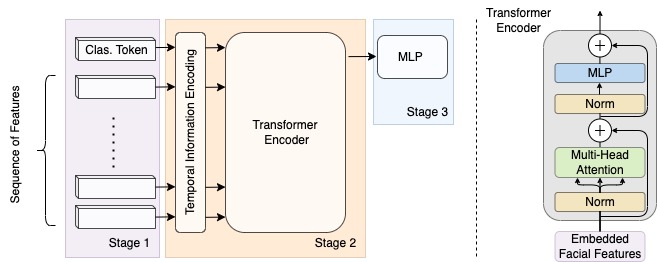}
    \caption{The model diagram of training the transformers.}
    \label{fig.model}
\end{figure*}
We implemented our models using Python 3.8.10 and PyTorch 1.12.0.0+cu102 and ran the experiments on the NVIDIA GeForce GTX 1080 GPU. The CAE model has a latent size of 128 with input and reconstructed images of size $96 \times 96$. The model was trained for 32 epochs using the Adam optimizer and Mean Square Error (MSE) loss function. The embedded facial features / latent feature vectors are updated until the MSE reaches a small value between the original face and the reconstructed one. 

The transformer was implemented in the same environment and trained for 40 epochs with Adam optimizer and weighted binary cross-entropy (BCE) loss function that is shown in Equation \ref{eq.WBCE}. The model consists of four transformer encoder layers, three positional embedding layers (sequential, sequence, and segment), and a classifier. The encoder has a hidden dimension equal to the latent feature vector size which is 128 and two heads. The classifier is a multilayer perceptron (MLP) which has three fully connected layers of [64, 32, 2]. The model is trained with a dropout of the value of 0.2. Figure \ref{fig.model} shows the transformer diagram. The first stage arranges the facial feature sequences with the appendant of a classification token. The second stage is adding the positional embeddings of the sequences and the transformer encoder. The third stage contains the MLP which takes the classification tokens output and makes a binary classification decision after a sigmoid function maps the model output.
\begin{equation} \label{eq.WBCE} 
Weighted BCE = - \beta (y_ilog(p_i) + (1-y_i)log(1-p_i))
\end{equation}
where $\beta$ is the weight assigned to the positive class. 

In our sequence approach, we defined $\textit{M} \in \mathbb{R^{lxF}}$ as a sequence where $l$ is the sequence size (15) and $F \in \mathbb{R^{128}}$ is the latent feature vector. The segments are defined as $S = \{M_1, M_2, ... M_s\} $. The video is presented as $V = \{ S_1, S_2, ... S_v\}$ or $\{M_1, M_2, ... M_n\}$. The positional embeddings are calculated as follows $P = P_M + P_S + P_p$. Eventually, the model input is $Z = z + P$ where z refers to the latent feature vectors. $P\in \mathbb{R^{128}}$ is the overall positional embeddings that capture the sequential representation of feature vectors because self-attention in transformers does not consider the timing appearance of the facial features. Considering the same concept, the positional embeddings within a video of the sequences and segments are captured in this layer. The purpose of this implementation is to have a hierarchical representation of every participant's video as we evaluate the model on the participants' video level.
\begin{table*}[H]
\caption{The model evaluation of 15 as a sequence size with positional embedding of sequence, segment, both, and none.}
\label{table.15_temporal_evaluation}
\resizebox{\textwidth}{!}{
\begin{tabular}{lcccccccccccc}
\hline
\multirow{3}{*}{Themes} & \multicolumn{12}{c}{positional information}                                                                                             \\
                        & \multicolumn{3}{c}{no position} & \multicolumn{3}{c}{sequence} & \multicolumn{3}{c}{segment} & \multicolumn{3}{c}{sequence and segment} \\ \cline{2-13} 
                        & Accuracy         & F1   & AUC   & Accuracy    & F1    & AUC    & Accuracy         & F1 & AUC & Accuracy             & F1      & AUC     \\ \cline{1-13} 
Summertime              & 76.7\%      & 0.77    & 0.77    & 76.6\%     & 0.79   & 0.76   & 80\%       & 0.81   & 0.8   & \textbf{83.3\%}         & 0.85       & 0.83       \\
Self-care               & \textbf{70\%}        & 0.64    & 0.69    & 63.3\%     & 0.56   & 0.63   & 60\%       & 0.5    & 0.59  & 66.7\%         & 0.58       & 0.66       \\
Halloween               & 84.4\%      & 0.86    & 0.84    & 75\%       & 0.79   & 0.74   & 81.3\%     & 0.84   & 0.8   & \textbf{87.5\%}         & 0.89       & 0.87       \\
Cities and Towns        & 76.9\%      & 0.77    & 0.77    & 76.9\%     & 0.76   & 0.77   & 79.5\%     & 0.79   & 0.8   & \textbf{79.5\%}         & 0.78       & 0.8       \\ \hline
\end{tabular}}
\end{table*}
\subsection{Results}
We separated the training and evaluation of the model based on the interview themes. Thus, all results will refer to the target theme. Accuracy, F1 score, and Area under the ROC Curve (AUC) are the evaluation metrics. The accuracy reflects the percentage of participants that were correctly classified during the testing. The F1 score is a combination score of precision and recall. AUC measures the model's ability to distinguish between binary cognitive conditions. We randomly separated the participants into 10-folds and performed cross-validation. We ensured the representation of participants from both cognitive conditions are assigned to the test fold during the random assigning. The procedure is to classify a video based on the percentage of the correct classified sequences within a video. We labeled a video as correctly classified if the majority of sequences within a video are correctly classified.

\begin{table}[H!]
\caption{Performance comparison with other MCI detection on I-CONECT dataset.}
\label{table.i-conect_studies}
\centering
\resizebox{\columnwidth}{!}{
\begin{tabular}{lcccc}
\hline
Methods                                           & Modality              & Accuracy    & F1      & AUC  \\ \hline
Asgari et al. (2017) \cite{asgari2017predicting} & Linguistic             & 83.3\%      & --      & 0.8  \\
Chen et al. (2020) \cite{chen2020topic}          &  Linguistic            &  79.2\%     & --      & 0.84  \\
Tang et al. (2022)   \cite{tang2022joint}       & Acoustic \& Linguistic  & --          & --      & 0.83  \\
Ours                                              & Visual                & 87.5\%      & 0.89    & 0.87 \\ \hline
\end{tabular}}
\end{table}
Table \ref{table.15_temporal_evaluation} shows the baseline performance of our model which considers the embedded facial features and does not consider the positions of the sequences and segments of the videos. The segments and sequences positional embeddings have improved the detection performance of the participants' cognitive conditions. The combinations of both of these positions performed better than either of them. In fact, applying one of these positions could drop the detection accuracy as it is presented for the Halloween theme when we add only one of these positional information the performance drops. The Self-care theme has the least detection performance among the themes which is impacted by the quality of the participants' facial appearance during the interview.

We can observe that by adding the segments and sequences information the transformer helped the model analyze the participants' interaction behavior or the length of their response to the interviewers' questions regardless of which theme is discussed. The results show that adding the segment and sequence information does improve the overall detection of the participants' condition. Thus, MCI is detectable using facial features; however, the interaction boosted the model performance.
\begin{table*}[hbt!]
\caption{The evaluation of different sizes of sequences.}
\label{table.different_seq_sizes}
\begin{tabular}{lccccccccc}
\hline
\multirow{3}{*}{Themes} & \multicolumn{9}{c}{sequence size}                                        \\
                        & \multicolumn{3}{c}{15} & \multicolumn{3}{c}{20} & \multicolumn{3}{c}{25} \\ \cline{2-10} 
                        & Accuracy & F1   & AUC  & Accuracy & F1   & AUC  & Accuracy & F1   & AUC  \\ \cline{1-10} 
Summertime              & \textbf{83.3\%}  & 0.85 & 0.83 & 80\%     & 0.82 & 0.79 & 77\%     & 0.8  & 0.76 \\
Self-care               & 66.7\%   & 0.58 & 0.66 & \textbf{70\%}     & 0.66 & 0.7  & 60\%     & 0.54 & 0.59 \\
Halloween               & \textbf{87.5\%}   & 0.89 & 0.87 & 81.2\%   & 0.83 & 0.81 & 81.2\%   & 0.84 & 0.8  \\
Cities and Towns        & \textbf{79.5\%}   & 0.78 & 0.8  & 79.5\%   & 0.78 & 0.8  & 79.5\%   & 0.76 & 0.8  \\ \hline
\end{tabular}
\end{table*}
Other studies have detected MCI participants in the I-CONECT dataset using different modalities. The main focus of the studies on the dataset was on speech signals. Table \ref{table.i-conect_studies} shows the best results of the studies and compares them with our top performance result among the themes in this study. Asgari \textit{et al.} \cite{asgari2017predicting} extracted Linguistic Inquiry and Word Count (LIWC) features from participants' transcribed data and classified the participants. Chen \textit{et al.} \cite{chen2020topic} have used a topic-based method to detect participants with MCI. Tang \textit{et al.} \cite{tang2022joint} integrated the acoustic and linguistic markers from the participants' speeches to distinguish the participants. 

\subsection{Ablation study}
In this section, we evaluated the model's detecting performance with variations of the model hyperparameters. We show the effects of different sequence sizes, overlapping of sequences, and loss function.  

\textbf{Sequence size}: we studied the effect of changing the sequence sizes when we feed the transformer with participants' facial features. Table \ref{table.different_seq_sizes} shows the performance of the model while applying different sequence sizes [15, 20, 25]. It is important to mention that changing the sequence size would change the total number of sequences per participant and the number of segments. In fact, the facial temporal features extracted from the transformer will be affected depending on the sequence size as well. The sequence size of 15 has performed overall better than the size of 20 and 25. The Cities and Towns theme performance was sustained across all sequence sizes. The prediction on the Self-care theme had a slight improvement when the sequence size is 20 but it dropped when the sequence size is 25.

\begin{table}[hbt!]
\caption{Detection comparison between sequence overlapping.}
\label{table.overlapping_results}
\centering
\resizebox{\columnwidth}{!}{
\begin{tabular}{lccc}
\hline
                 & \multicolumn{3}{c}{sequence overlapping percentage} \\
Themes           & 0\%             & 20\%            & 40\%            \\ \hline
Summertime       & \textbf{83.3\%}          & 80\%            & 80\%            \\
Self-care        & \textbf{66.7\%}          & 63.3\%          & 60\%            \\
Halloween        & \textbf{87.5\%}          & 84.4\%          & 75\%            \\
Cities and Towns & 79.5\%          & 79.5\%          & \textbf{82.1\%}          \\ \hline

\end{tabular}}
\end{table}
\textbf{Sequence overlapping}: we studied the effect of sequence overlapping while keeping the sequence size fixed. Table \ref{table.overlapping_results} shows the accuracies of 0\%, 20\%, and 40\% overlapping sizes while selecting sequences of facial features during the training and evaluation. This study shows our sequence covered the majority of participants' features during the interviews. The overlapping of sequences will increase the number of sequences and keep the number of segments fixed. Thus, the accuracies decreased with the overlapping except for the Cities and Towns themes. With the consideration of Cities and Towns theme performance shown in Table \ref{table.15_temporal_evaluation}, the accuracy of adding segment information was similar to adding sequence and segment information. However, the increase of sequence information caused the model a better prediction of the participants' cognitive conditions. In other words, sequence augmentation improved the prediction for the Cities and Towns theme. 

\begin{table}[h!]
\caption{Comparison between applying weighted BCE and BCE as a model loss function.}
\label{table.loss_function_comparison}
\centering
\begin{tabular}{lcccccc}
\hline
Themes           & weighted BCE & BCE \\ \hline
Summertime       & \textbf{83.3\%}         & 76.7\%     \\
Self-care        & \textbf{66.7\%}        & 56.7\%     \\
Halloween        & \textbf{87.5\%}         & 78.1\%    \\
Cities and Towns & \textbf{79.5\%}          & 76.9\%     \\ \hline
\end{tabular}
\end{table}
\textbf{Loss function}: although participants’ conditions are balanced for all selected themes, the number of sequences that represent each participant varies so the weighted binary cross-entropy loss function triggers the class imbalance in the training for every fold. Here, we study the concept of applying the weighted BCE loss function to help the model performance compared with the average BCE loss function \ref{eq.BCE}. Table \ref{table.loss_function_comparison} shows the accuracies of different themes with consideration of changing the loss function. Across all themes, the accuracies improved by at least 2.6\% with the weighted BCE loss function. We investigate this to prove that the model prediction performance improved with the weighted BCE loss function. This loss function would act as if the class with fewer samples has more samples based on the ratio weight that is updated for every fold during the training.
\begin{equation} \label{eq.BCE} 
BCE = - \frac{1}{N} \sum_{i=1}^{N} (y_ilog(p_i) + (1-y_i)log(1-p_i))
\end{equation}

Our work has several limitations, especially regarding the video quality. In our approach, the manual quality check is one of the challenges which takes a considerably long time to evaluate. However, an automated computer-based approach will help achieve the evaluation faster; therefore, more themes can be included in the study. The number of eliminated videos within a theme is concernedly large which is mainly based on the video quality. The video quality does not only refer to the typical technical challenges of the videos such as room lighting, viewpoint, and facial appearance but it also considers the full face during the interview because some participants did not show their faces for the majority of the video. Therefore, older adults are having challenges in handling computer intermedium interviews.
}

\section{Conclusion} \label{sec.conclusion}
In this paper, we demonstrated a method to detect older adults with MCI using their facial features in visual semi-structured interviews from the I-CONECT study. The method is based on extracting spatial features and temporal information that is conducted by using a CAE and the self-attention mechanism in the transformers. We showed the interaction timing of the facial features of subjects is important as it helped boost the performance of our model in distinguishing the cognitive condition among participants. We conclude that using segments and sequence indices within a video does improve the prediction of the participants' conditions. Participants’ facial features vary during the topic discussion. Thus, future work will study the sequences' contribution to cognitive condition detection among the participants including the speech data and the automated video quality assessment using DL algorithms.

\printcredits

\section*{Declaration of Competing Interest}
The authors declare that they have no known competing financial interests or personal relationships that could have appeared to influence the work reported in this paper.

\section*{Acknowledgements}
The following federal fund supported this project: R01AG051628, R01AG056102, RF1AG672448 from the National Institute of Health (NIH) in the United States to Dr. Hiroko Dodge. Dr. Mohammad Mahoor received a grant from the Colorado Office of Economic Development and International Trade in support of this research project.

\bibliographystyle{elsarticle-num}

\bibliography{main}



\end{document}